\documentclass{article}

\usepackage{arxiv}

\usepackage[utf8]{inputenc} 
\usepackage[T1]{fontenc}    
\usepackage{hyperref}       
\usepackage{url}            
\usepackage{booktabs}       
\usepackage{amsfonts}       
\usepackage{nicefrac}       
\usepackage{microtype}      
\usepackage{lipsum}
\usepackage{graphicx}

\usepackage{cite}
\usepackage{amsmath,amssymb,amsfonts}
\usepackage{algorithmic}
\usepackage{graphicx,subcaption}
\usepackage{textcomp}
\usepackage{xcolor}
\graphicspath{ {./images/} }

\title{Complexity Controlled Generative Adversarial Networks}

\author{
 Himanshu Pant, Jayadeva and Sumit Soman \\
  Department of Electrical Engineering\\
  Indian Institute of Technology\\
  Delhi, India \\
  \texttt{jayadeva@ee.iitd.ac.in} \\
}

\begin{document}
\maketitle
\begin{abstract}
One of the issues faced in training Generative Adversarial Nets (GANs) and their variants is the problem of mode collapse, wherein the training stability in terms of the generative loss increases as more training data is used. In this paper, we propose an alternative architecture via the Low-Complexity Neural Network (LCNN), which attempts to learn models with low complexity. The motivation is that controlling model complexity leads to models that do not overfit the training data. We incorporate the LCNN loss function for GANs, Deep Convolutional GANs (DCGANs) and Spectral Normalized GANs (SNGANs), in order to develop hybrid architectures called the LCNN-GAN, LCNN-DCGAN and LCNN-SNGAN respectively. On various large benchmark image datasets, we show that the use of our proposed models results in stable training while avoiding the problem of mode collapse, resulting in better training stability. We also show how the learning behavior can be controlled by a hyperparameter in the LCNN functional, which also provides an improved inception score.
\end{abstract}


\section{Introduction}

Generative Adversarial Networks (GANs) have recently emerged as an interesting area of research in the deep learning domain. GANs conventionally comprise of a generator and a discriminator. The generator attempts to continually generate fake images, while the discriminator network aims to differentiate between real and fake images. As the network is trained, it is expected that the ability of the generator and discriminator in their respective tasks would both improve, based on the concept of a \textit{``zero sum game.''}

In terms of learning theory, Generative Adversarial Networks (GANs) \cite{goodfellow2014generative} are a class of learning models that involve the training of a generative model $G$ and a discriminative model $D$. $G$ tries to capture the generator distribution and $D$ tries to estimate whether a training sample belongs to the data, rather than the distribution modeled by $G$. Specifically, given a prior $p_z (z)$ on input noise variable and data mapping $P(z;\phi_G)$ and $P(x;\phi_D)$ for the generator and discriminator models (which are conventionally Multi-Layer Perceptrons), GAN attempts to solve the optimization problem given by
\begin{gather}
    \min_G \max_D \mathbb{E}_{x\sim p_{data}(x)} log[D(x)] +  \mathbb{E}_{z\sim p_{z}(z)} log[1-D(G(z))] \label{eqn:gan_obj}
\end{gather}

GANs have tried to overcome the limitations faced by techniques such as Maximum Likelihood Estimation (MLE) in the computation of probabilities (which tend to become intractable). However, they face issues in training stability, and restrictions in  visualization when compared to techniques such as Convolutional Neural Networks (CNNs).

Other variants of GANs include Least Squared GAN \cite{mao2017least}, which uses a least squared loss function for the discriminator in order to address the vanishing gradient problem faced with conventional GANs. LS-GANs show improved image quality and better training stability compared to the baseline. Variants have also been developed using Integral Probability Metrics (IPM) such as Mean and Covariance Feature Matching GAN \cite{mroueh2017mcgan} and Fisher GAN \cite{mroueh2017fisher}. Other approaches include Maximum Mean Discrepancy (MMD) \cite{li2017mmd, li2015generative}, margin adaptation \cite{wang2017magan}, Boundary Equilibrium GAN \cite{berthelot2017began}, Energy-Based GAN \cite{zhao2016energy}, probabilistic GAN \cite{eghbal2017probabilistic}, Bayesian GAN \cite{saatci2017bayesian}, conditional GAN \cite{mirza2014conditional}, Wasserstein GAN \cite{arjovsky2017wasserstein, bellemare2017cramer}, among others.

Deep Convolutional GANs (DCGANs) were proposed \cite{radford2015unsupervised} as an approach to implement GANs with Convolutional Neural Networks (CNNs) for unsupervised image classification. The convolutional layers implemented in the generator network has been shown to give better results than conventional GANs. The network proposed in \cite{radford2015unsupervised} applies convolutional layers to reshape the network via a `project and reshape' approach and the same has been shown to be beneficial for large-scale scene generation. The use of the CNN-based approach has also been shown to have improved training stability resulting in optimally trained discriminator networks while also facilitating visualization via filters.

One of the primary challenges elucidated in the concluding section of \cite{radford2015unsupervised} is the problem of \textit{``collapse of a subset of filters to a single oscillating node''} over prolonged training, which is termed as \textit{mode collapse}. In this paper, we discuss a mechanism for addressing this problem in GANs and DCGANs, motivated by an approach for reducing model complexity in neural networks.

Spectral Normalized GANs (SN-GANs) \cite{miyato2018spectral} introduce a  normalization scheme that uses a single hyperparameter in order to prevents the discriminator from being unstable during training, particularly in high dimensional spaces. The optimal discriminator is given by
\begin{gather}
D^*_G (x)=\text{sigmoid}(f^*(x))
\end{gather}

\noindent where
\begin{gather}
f^*(x)=log(q_{data}(x)-log (p_G(x)))
\end{gather}

The spectral normalization scheme adopted by the SN-GAN normalizes the weight matrix $W$ so that its Lipschitz constant $\sigma(W)=1$, which is realized as
\begin{gather}
\hat{W}_{SN}(W)=\frac{W}{\sigma(W)}
\end{gather}

The computation of the spectral norm is made computationally efficient in the case of SN-GAN by using the power iteration method. SN-GANs have been shown to give improved results compared to other stabilization and weight normalization techniques in literature.


\section{Low Complexity Neural Network}

The LCNN \cite{pant2017learning} aims to introduce the notion of controlling model complexity in a neural network by modifying the cost function. Specifically, we minimize the error with explicit $L_2$ regularizers on weights along with the model complexity term. 

Consider a family of large margin neural network classifiers in which each has $k$ layers and one output neuron, where all neurons have real valued weights and a constant bias input each. All neurons use a continuous, monotonically non-decreasing activation function. The class of any input sample is determined by thresholding the output of the neuron in the final layer keeping the weights in previous layers fixed. Let the number of neurons in the penultimate layer be $n$. Let the margin of the classifier obtained by thresholding the output neuron be at least $d_{min} > 0$. Then, the VC dimension $\gamma$ of this neural network satisfies

\begin{equation}\label{eqnha}
  \gamma \leq 1 + \operatorname{Min} \left( \frac{4R^2}{d_{min}^2}, n \right)
\end{equation}
where $R$ denotes the radius of the smallest sphere enclosing $\left\{ V^i_{(k - 1), 1}, V^i_{(k - 1), 2},..., V^i_{(k - 1), n} \right\}$, $i = 1, 2, ..., M$.

The set of input samples $x^i$, $i = 1, 2, ..., M$, has a finite radius $R_{input} < \infty$. The class of any input sample is determined by thresholding the output of the neuron in the final layer. The outputs of the penultimate layer neurons when the $i-th$ sample is presented at the input, are denoted by $\{ V^i_{(k - 1), 1}, V^i_{(k - 1), 2},..., V^i_{(k - 1), n}\}$. The weight vector of the output neuron is denoted by $w$, and its bias input is denoted by $b$. In other words, when the $i-th$ training sample is presented at the input, the net input to the final layer neuron is given by
\begin{gather}
  net^i = \sum_{j = 1}^n w_j ~ V^i_{(k-1),j} \;+\; b =  \mathbf{w}^T \mathbf{V}^i_{(k-1)} + b,
\end{gather}
and the class predicted by the classifier is determined from the sign of $net^i$. Further, assume that any member of the family of classifiers so defined has a margin of at least $d_{min}$, where $d_{min}$ is strictly positive. Then, the VC dimension $\gamma$ of this classifier is bounded from above by
\begin{gather}
  \gamma \leq 1 + \operatorname{Min} \left( C\sum_{i=1}^M (net^i)^2 \text{ , } \; n \right), \label{ubound1}
\end{gather}
where $C$ is a real positive constant.


For binary classification with labels $y_i \in \{-1,1\}$  or $y_i \in \{0,1\}$, the empirical error $E_{emp}$ is given as
		 \begin{gather}
			E_{emp} =  \left[ \frac{1}{M} \sum_{i=1}^M \left( -log(\operatorname{softmax}(net^i)) \right)\right]
	      \end{gather}
	      where $M$ denotes the number of training samples and $f(net^i)$ denotes the activation function $f(\cdot)$ on net input to neuron $i$ given by $net^i$.

In regularized neural networks, this is modified by adding a term proportional to $\|w\|^2$ at individual weights (weight decay). In order to minimize model complexity, the Low Complexity Neural Network (LCNN) classifier uses the modified error functional $E$ of the form

\begin{gather}
	 E =  E_{emp}+ C~\sum_{i=1}^M (net^i)^2
	\end{gather}

\section{GANs with LCNN}

As mentioned in the Introduction, the generator and discriminator in a GAN compete for improved performance. The generator loss determines the performance of the generator, while the discriminator loss is indicative of the error being made in distinguishing the real and adversarial examples. One of the primary reasons for mode collapse and unstable training is the divergence of the generator and discriminator losses over the course of their training. Our approach, based on the LCNN, involves introducing an additional term in the objective function of the GAN which addresses the issue of mode collapse. The motivation is that controlling model complexity leads to models that do not overfit the training data.

We add the LCNN loss functional in the final layer of the generator and discriminator which allows for minimization of the activations of the neurons in the  final layer, driving them closer to the origin. During the course of training, when the difference of the error between the discriminator and generator diverges, the LCNN term ($\min \|w^T x + b\|$) tries to maintain the balance between discriminator and generator errors by pushing activations generated by both real and generated samples towards minimum, hence making it difficult for the discriminator to distinguish between real and generated samples. This improves stability and the quality of generated images as more and more data are presented to the GAN. Also, since we conventionally use the tan-sigmoid or log-sigmoid activation functions, and these have maximum gradient near the origin, the vanishing gradient problem is mitigated during backpropagation, which also results in improved training stability.

\subsection{LCNN-GAN}

The modified objective function for the LCNN-GAN is obtained by adding an additional term to the conventional objective function of GANs which can be represented as

\begin{gather}
    \min_G \max_D \mathbb{E}_{x\sim p_{data}(x)} log[D(x)] +  \mathbb{E}_{z\sim p_{z}(z)} log[1-D(G(z))] + \nonumber \\
    \min_D (C_1 L_C(D(x))+C_2 L_C(D(G(x)))) \label{eqn:lcnn_gan_obj}
\end{gather}

where $L_C$ is the LCNN term applied on the last layer of $D$. Specifically, $L_C=\sum_{i=1}^M (net^i)^2$ for the discriminator. Here $C_1$ and $C_2$  are hyper-parameters that can be tuned by the user. We call the model using the objective function above as the LCNN-GAN.

Similar analogy can also be extended for obtaining the hybrid LCNN-incorporated  objective functions for the DC-GANs and for SN-GANs in order to obtain the LCNN-DCGAN and LCNN-SNGAN respectively. The final layers in these hybrid models use the softmax activation function with cross entropy loss.

\section{Experiments}


We now present results on the LCNN incorporated approaches for GANs, DC-GANs and SN-GANs in the following subsections. The conventional architectures  proposed in literature for these GAN models have been used in our experiments. The results are presented in terms of generator and discriminator loss, as well as in terms of representative images. Subsequently, we also present results in terms of inception score for the various LCNN-incorporated GAN models discussed in this paper. For these experiments, the values of hyper-parameters have been set as the same for both $C_1$ and $C_2$ and have been indicated as $C$ in the plots comparing the generator and discriminator loss.

\subsection{Results on LCNN-GAN}

\begin{figure}[h]
\centering
        \begin{subfigure}[b]{0.5\textwidth}
            \includegraphics[width=\linewidth]{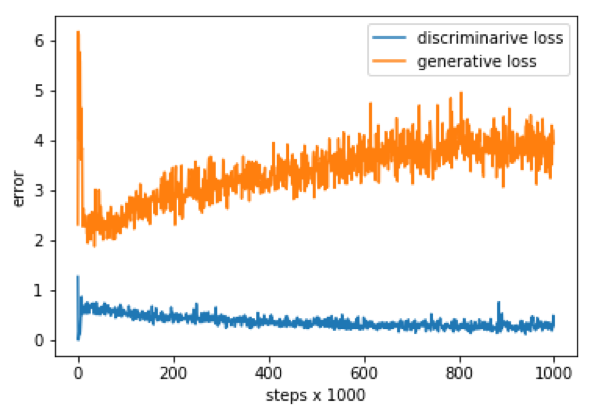}
            \caption{Mode Collapse in GAN}
            \label{fig:mode_gan}
        \end{subfigure}%
        \begin{subfigure}[b]{0.5\textwidth}
            \includegraphics[width=\linewidth]{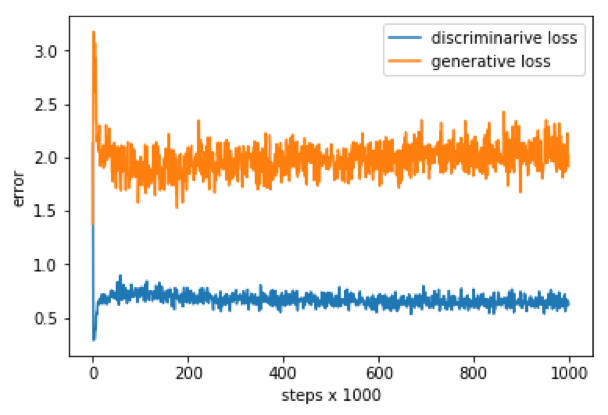}
            \caption{LCNN-GAN}
            \label{fig:mode_gan_lcnn}
        \end{subfigure}%
        \caption{Mode Collapse Analysis for MNIST Data}
        \label{fig:1}
\end{figure}

In Fig. \ref{fig:1} we compare mode collapse for GAN in Fig. \ref{fig:mode_gan} and the case for GAN with LCNN in Fig. \ref{fig:mode_gan_lcnn} for the MNIST dataset. It can be seen that the generative loss is contained in case of the LCNN-GAN, as opposed to the conventional GAN, thereby avoiding mode collapse.

\begin{figure*}
\centering
        \begin{subfigure}[b]{0.15\textwidth}
            \includegraphics[width=\linewidth]{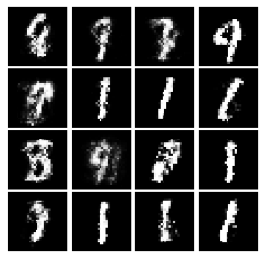}
            \caption{10k}
            \label{fig:gan_10k}
        \end{subfigure}%
        \begin{subfigure}[b]{0.15\textwidth}
            \includegraphics[width=\linewidth]{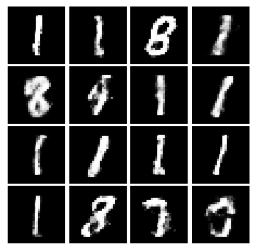}
            \caption{50k}
            \label{fig:gan_50k}
        \end{subfigure}%
        \begin{subfigure}[b]{0.15\textwidth}
            \includegraphics[width=\linewidth]{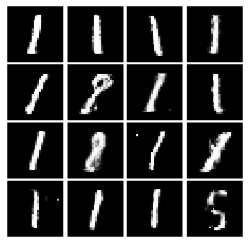}
            \caption{100k}
            \label{fig:gan_100k}
        \end{subfigure}%
        \begin{subfigure}[b]{0.15\textwidth}
            \includegraphics[width=\linewidth]{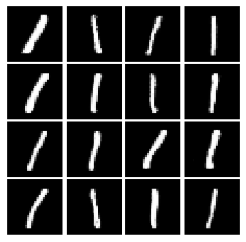}
            \caption{200k}
            \label{fig:gan_200k}
        \end{subfigure}%
    \caption{GAN Generated Fake MNIST Samples at different training stages}
    \label{fig:3}
\end{figure*}

\begin{figure*}
\centering
        \begin{subfigure}[b]{0.15\textwidth}
            \includegraphics[width=\linewidth]{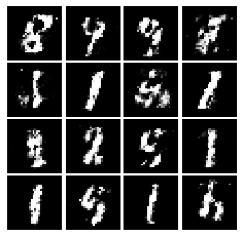}
            \caption{10k}
            \label{fig:lcnn_gan_10k}
        \end{subfigure}%
        \begin{subfigure}[b]{0.15\textwidth}
            \includegraphics[width=\linewidth]{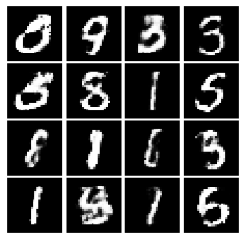}
            \caption{50k}
            \label{fig:lcnn_gan_50k}
        \end{subfigure}%
        \begin{subfigure}[b]{0.15\textwidth}
            \includegraphics[width=\linewidth]{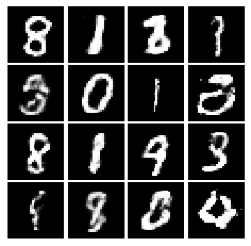}
            \caption{100k}
            \label{fig:lcnn_gan_100k}
        \end{subfigure}%
        \begin{subfigure}[b]{0.15\textwidth}
            \includegraphics[width=\linewidth]{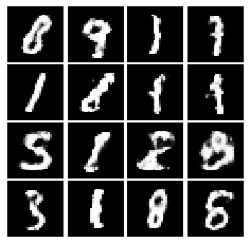}
            \caption{200k}
            \label{fig:lcnn_gan_200k}
        \end{subfigure}%
        \begin{subfigure}[b]{0.15\textwidth}
            \includegraphics[width=\linewidth]{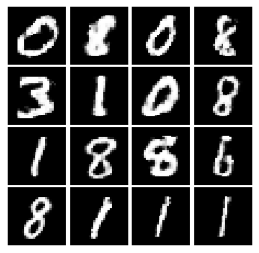}
            \caption{500k}
            \label{fig:lcnn_gan_500k}
        \end{subfigure}%
        \begin{subfigure}[b]{0.15\textwidth}
            \includegraphics[width=\linewidth]{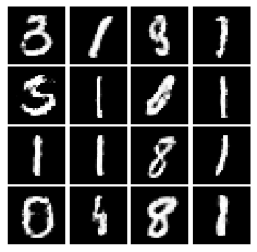}
            \caption{1000k}
            \label{fig:lcnn_gan_1000k}
        \end{subfigure}%
        \caption{GAN-LCNN Generated Fake MNIST Samples at different training stages (numbers indicate mini batches processed)}
        \label{fig:2}
\end{figure*}

We also show the fake images generated by the GAN and LCNN-GAN to illustrate how mode collapse is avoided in case of LCNN-GAN. As can be seen from Fig. \ref{fig:2} for the LCNN-GAN, the images generated as the number of mini batches are increased (Figs. \ref{fig:lcnn_gan_10k} - \ref{fig:lcnn_gan_1000k}) retains the characteristics of the individual digits. However, in case of the conventional GAN, with an increase in the number of mini batches processed, (Figs. \ref{fig:gan_10k} - \ref{fig:gan_200k}), mode collapse results in all images being similar to the digit 1. This illustrates that mode collapse is avoided in case of the LCNN-GAN.

\begin{figure}[h]
\centering
        \begin{subfigure}[b]{0.5\textwidth}
            \includegraphics[width=\linewidth]{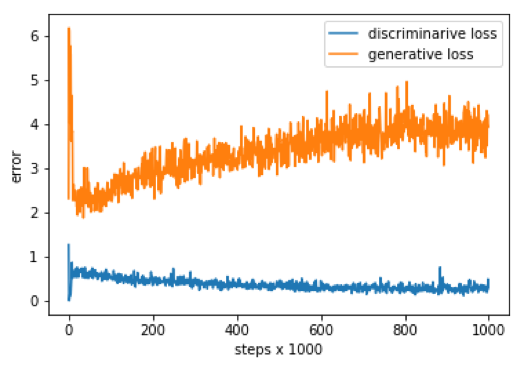}
            \caption{GAN}
            \label{fig:c001_gan}
        \end{subfigure}%
        \begin{subfigure}[b]{0.5\textwidth}
            \includegraphics[width=\linewidth]{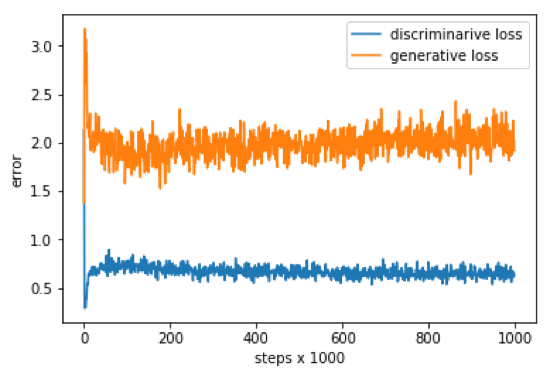}
            \caption{LCNN-GAN}
            \label{fig:c001_lcnn_gan}
        \end{subfigure}%
        \caption{Balance between generator and discriminator for C=0.01}
        \label{fig:4}
\end{figure}

We discuss how the hyperparameter C in the LCNN-GAN affects balance between the generative and discriminative loss. In Fig. \ref{fig:4} we compare the GAN and LCNN-GAN for C=0.01. It can be seen that the loss is contained in case of LCNN-GAN.

\begin{figure}[h]
\centering
        \begin{subfigure}[b]{0.45\textwidth}
            \includegraphics[width=\linewidth]{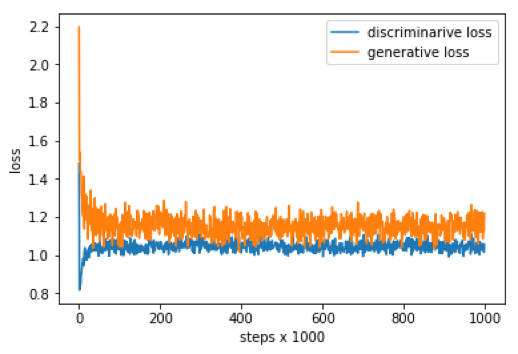}
            \caption{C=0.1}
            \label{fig:gan_bal_lcnn_c0_1}
        \end{subfigure}%
        ~
        \begin{subfigure}[b]{0.3\textwidth}
            \includegraphics[width=\linewidth]{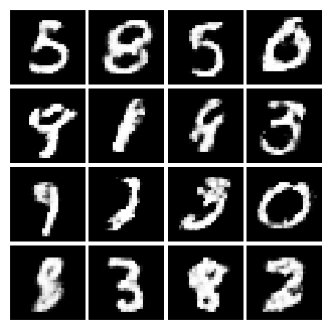}
            \caption{C=0.1}
            \label{fig:gan_bal_lcnn_mnist_c0_1}
        \end{subfigure}

        \begin{subfigure}[b]{0.45\textwidth}
            \includegraphics[width=\linewidth]{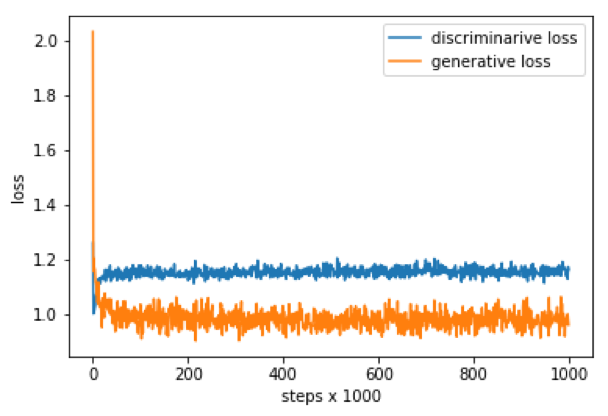}
            \caption{C=0.2}
            \label{fig:gan_bal_lcnn_c0_2}
        \end{subfigure}%
        ~
        \begin{subfigure}[b]{0.3\textwidth}
            \includegraphics[width=\linewidth]{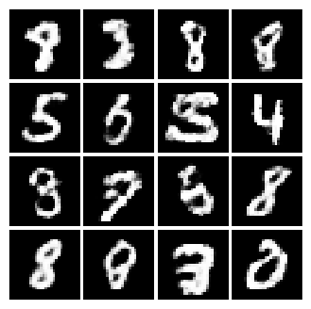}
            \caption{C=0.2}
            \label{fig:gan_bal_lcnn_mnist_c0_2}
        \end{subfigure}
        \caption{Balance between generator and discriminator for different values of C for LCNN-GAN}
        \label{fig:5}
\end{figure}

The effect of hyperparameter C can also be seen in Fig. \ref{fig:5}, where we compare the loss for C=0.1 and C=0.2 in Figs. \ref{fig:gan_bal_lcnn_c0_1} and \ref{fig:gan_bal_lcnn_c0_2} respectively. Representative images from the MNIST dataset are also shown alongside in Figs. \ref{fig:gan_bal_lcnn_mnist_c0_1} and \ref{fig:gan_bal_lcnn_mnist_c0_2} respectively. It can be seen that the hyperparameter $C$ plays an important role in balancing the error between generator and discriminator. Also, by varying $C$ during training we can control generator and discriminator behavior.

Since MNIST has 10 classes, most of the energy (variance) content of MNIST will be encompassed by top 10 eigenvalues. On doing PCA on MNIST train data set it was found that top 10 eigenvalues correspond to 83\% energy. For analyzing variance and modes generated in GAN and LCNN-GAN
in each iteration, we calculated how many eigenvectors correspond to 83\% energy content for GAN and LCNN-GAN. In a way, we are deducing that how many modes GAN and LCNN-GAN are capable of generating in fake images.

\begin{figure}
\centering
        \begin{subfigure}[b]{0.5\textwidth}
            \includegraphics[width=\linewidth]{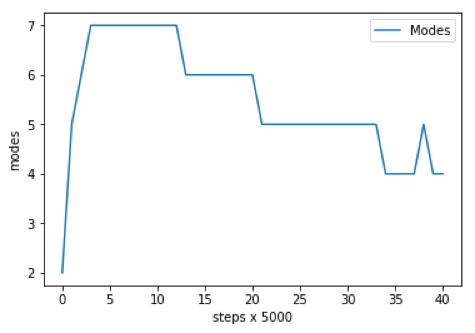}
            \caption{GAN}
            \label{fig:mode_gan_mnist}
        \end{subfigure}%
        \begin{subfigure}[b]{0.5\textwidth}
            \includegraphics[width=\linewidth]{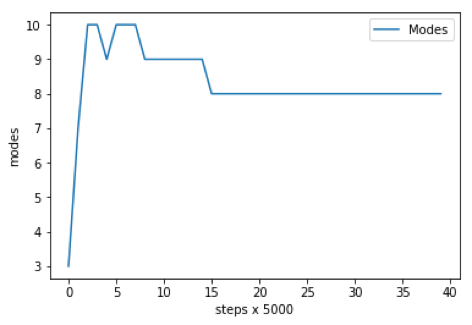}
            \caption{LCNN-GAN}
            \label{fig:mode_gan_lcnn_mnist}
        \end{subfigure}%
        \caption{Energy Content and Modes study for GAN and LCNN-GAN}
        \label{fig:6}
\end{figure}

Results are shown in Fig.\ref{fig:6}, for GAN in Fig. \ref{fig:mode_gan_mnist} and LCNN-GAN in Fig. \ref{fig:mode_gan_lcnn_mnist}. It can be seen that the LCNN-GAN retains many more modes when compared to GAN, providing further evidence in favor of addressing the mode collapse problem faced in GANs.

\subsection{Results on LCNN-DCGAN}

\subsubsection{Dataset Used}
CelebFaces Attributes Dataset (CelebA) is a large-scale face attributes dataset with more than 200K celebrity images, each with 40 attribute annotations. The images in this dataset cover large pose variations and background clutter. CelebA has large diversities, large quantities, and rich annotations, including
\begin{enumerate}
    \item 10,177 identities
    \item 202,599 face images
    \item 5 landmark locations, 40 binary attributes annotations per image.
\end{enumerate}

\begin{figure}
\centering

        \begin{subfigure}[b]{0.49\textwidth}
            \includegraphics[width=\linewidth]{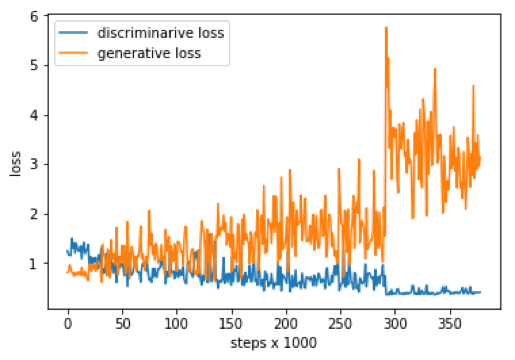}
            \caption{DCGAN}
            \label{fig:dcgan_stable}
        \end{subfigure}%
         \begin{subfigure}[b]{0.49\textwidth}
            \includegraphics[width=\linewidth]{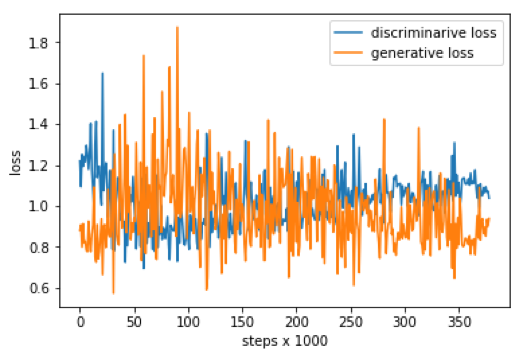}
            \caption{LCNN DCGAN}
            \label{fig:dcgan_lcnn_stable}
        \end{subfigure}%
        \caption{Training Stability for DCGAN and LCNN-DCGAN}
        \label{fig:7}
\end{figure}

In Fig. \ref{fig:7} we show the training stability for the DCGAN and LCNN-DCGAN in Figs. \ref{fig:dcgan_stable} and \ref{fig:dcgan_lcnn_stable} respectively. It can be seen that in case of the DCGAN, the generative and discriminative losses tend to rapidly vary, while it is contained in case of the LCNN-DCGAN. It may be noted here that the values of hyperparameter C have varied linearly from 0.01 to 0.1 for the LCNN-DCGAN plot.

\begin{figure}
\centering
        \begin{subfigure}[b]{0.25\textwidth}
        \centering
            \includegraphics[scale=0.38]{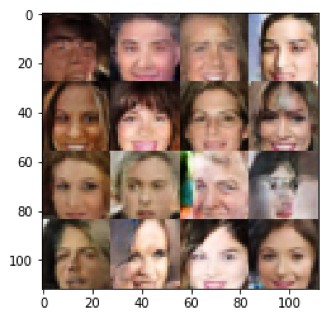}
            \caption{DCGAN 1 (Before)}
            \label{fig:dcgan_before_1}
        \end{subfigure}%
        \begin{subfigure}[b]{0.25\textwidth}
        \centering
            \includegraphics[scale=0.38]{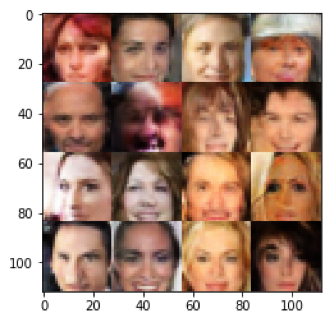}
            \caption{DCGAN 2 (Before)}
            \label{fig:dcgan_before_2}
        \end{subfigure}%
        \begin{subfigure}[b]{0.25\textwidth}
        \centering
            \includegraphics[scale=0.38]{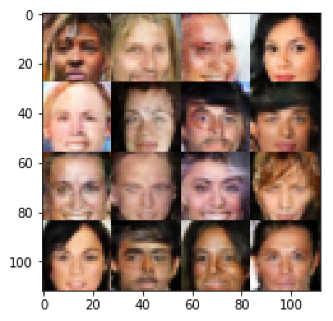}
            \caption{DCGAN 3 (Before)}
            \label{fig:dcgan_before_3}
        \end{subfigure}%

        \begin{subfigure}[b]{0.25\textwidth}
        \centering
            \includegraphics[scale=0.38]{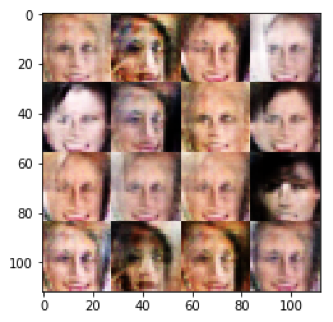}
            \caption{DCGAN 1 (After)}
            \label{fig:dcgan_after_1}
        \end{subfigure}%
        \begin{subfigure}[b]{0.25\textwidth}
        \centering
            \includegraphics[scale=0.38]{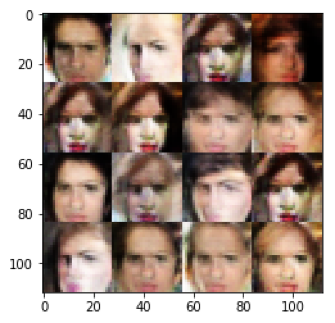}
            \caption{DCGAN 2 (After)}
            \label{fig:dcgan_after_2}
        \end{subfigure}%
        \begin{subfigure}[b]{0.25\textwidth}
        \centering
            \includegraphics[scale=0.38]{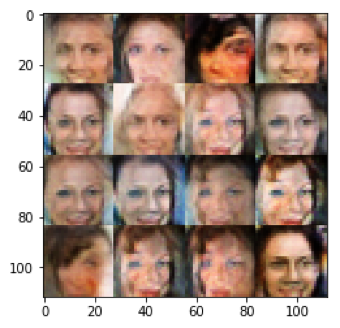}
            \caption{DCGAN 3 (After)}
            \label{fig:dcgan_after_3}
        \end{subfigure}%

        \begin{subfigure}[b]{0.25\textwidth}
        \centering
            \includegraphics[scale=0.38]{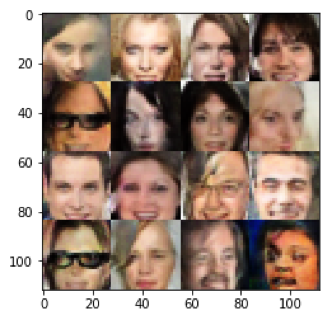}
            \caption{LCNN-DCGAN 1}
            \label{fig:dcgan_lcnn_1}
        \end{subfigure}%
        \begin{subfigure}[b]{0.25\textwidth}
        \centering
            \includegraphics[scale=0.38]{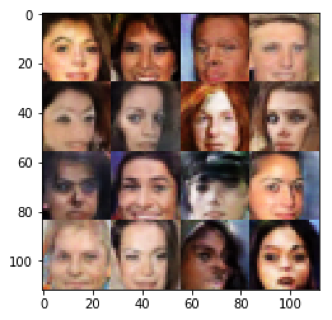}
            \caption{LCNN-DCGAN 2}
            \label{fig:dcgan_lcnn_2}
        \end{subfigure}%
        \begin{subfigure}[b]{0.25\textwidth}
        \centering
            \includegraphics[scale=0.38]{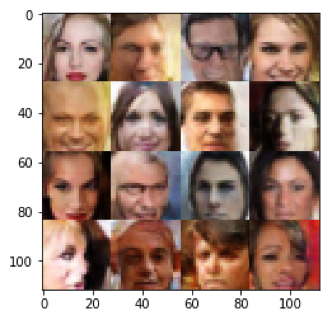}
            \caption{LCNN-DCGAN 3}
            \label{fig:dcgan_lcnn_3}
        \end{subfigure}%
\caption{Fake Images Generated from DCGAN before and after Unstable/Mode Collapse compared with LCNN-DCGAN}
\label{fig:8}
\end{figure}

In Fig. \ref{fig:8}, we show representative images from the CelebA dataset for the case of DCGAN before (Figs. \ref{fig:dcgan_before_1}-\ref{fig:dcgan_before_3}) and after (Figs. \ref{fig:dcgan_after_1}-\ref{fig:dcgan_after_3}) mode collapse, and compare them with those obtained from the LCNN-DCGAN (Figs. \ref{fig:dcgan_lcnn_1}-\ref{fig:dcgan_lcnn_3}). It can be seen that the mode collapse occurring in case of DCGANs is controlled in case of using the LCNN-DCGAN.

\subsection{CIFAR-10}

\begin{figure}
    \centering
    \includegraphics[scale=0.8]{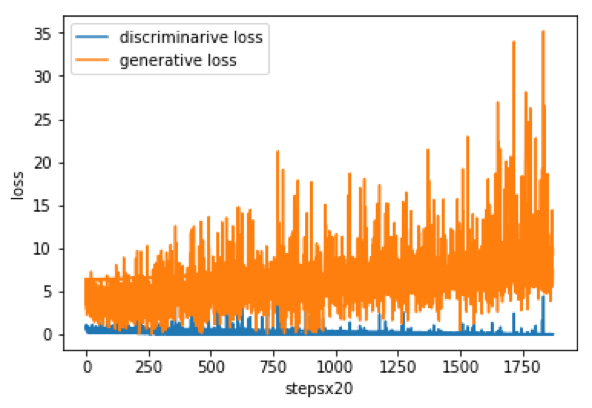}
    \caption{Generative and Discriminative Loss for DCGAN on CIFAR-10}
    \label{fig:9}
\end{figure}

As a baseline, we show the generative and discriminative loss for the DCGAN on the CIFAR-10 dataset in Fig. \ref{fig:9}.

\begin{figure*}
\centering
        \begin{subfigure}[b]{0.29\textwidth}
            \includegraphics[width=\linewidth]{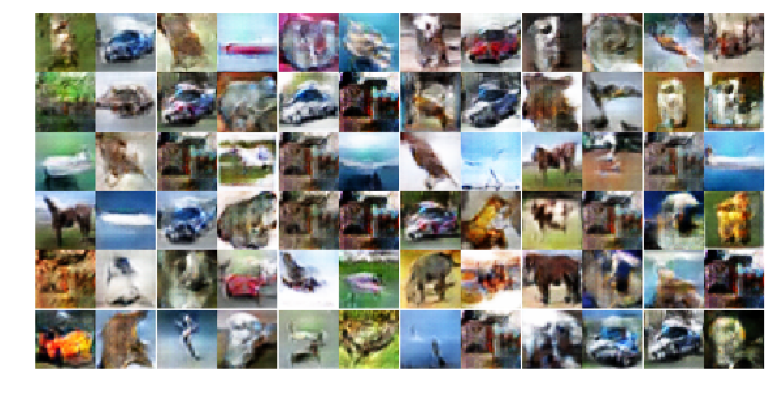}
            \caption{DCGAN 1 (Before)}
            \label{fig:dcgan_cifar_before_1}
        \end{subfigure}%
        \begin{subfigure}[b]{0.29\textwidth}
            \includegraphics[width=\linewidth]{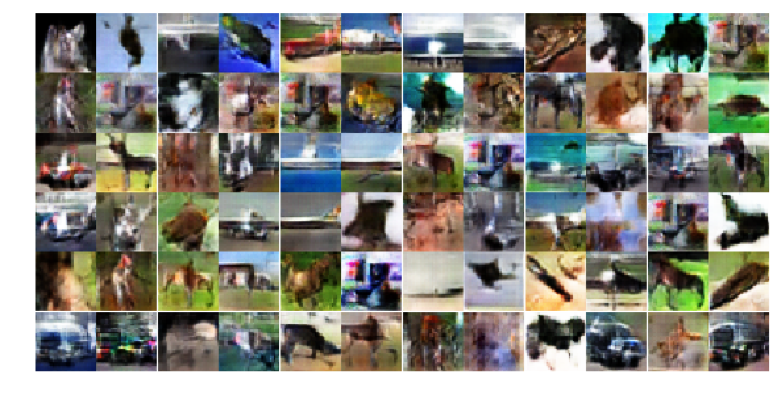}
            \caption{DCGAN 2 (Before)}
            \label{fig:dcgan_cifar_before_2}
        \end{subfigure}%

        \begin{subfigure}[b]{0.29\textwidth}
            \includegraphics[width=\linewidth]{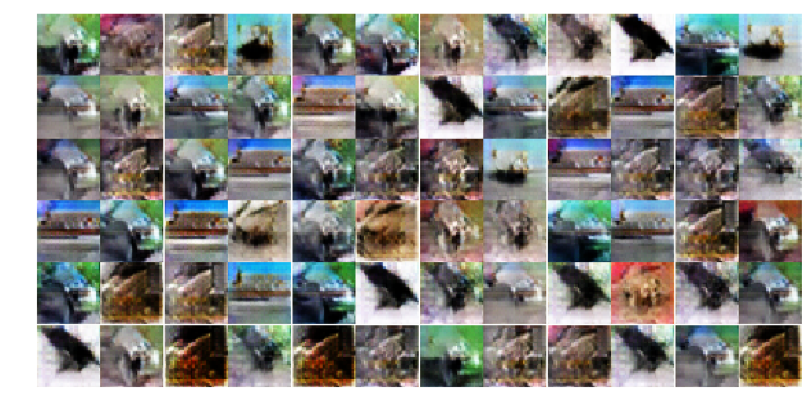}
            \caption{DCGAN 1 (After)}
            \label{fig:dcgan_cifar_after_1}
        \end{subfigure}%
        \begin{subfigure}[b]{0.29\textwidth}
            \includegraphics[width=\linewidth]{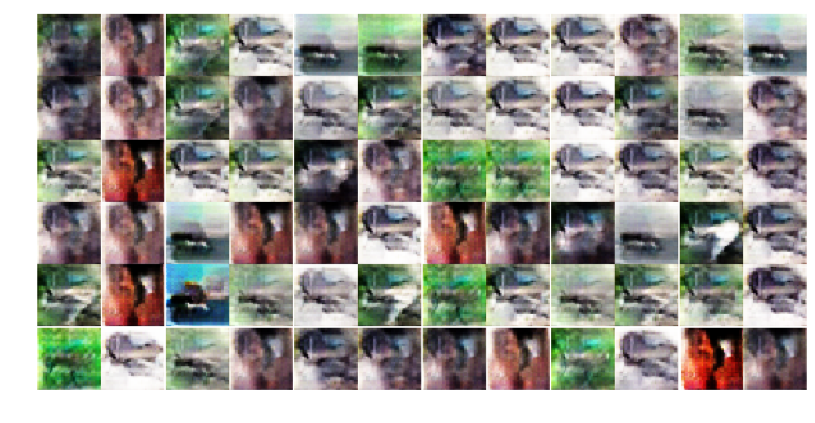}
            \caption{DCGAN 2 (After)}
            \label{fig:dcgan_cifar_after_2}
        \end{subfigure}%
        \begin{subfigure}[b]{0.29\textwidth}
            \includegraphics[width=\linewidth]{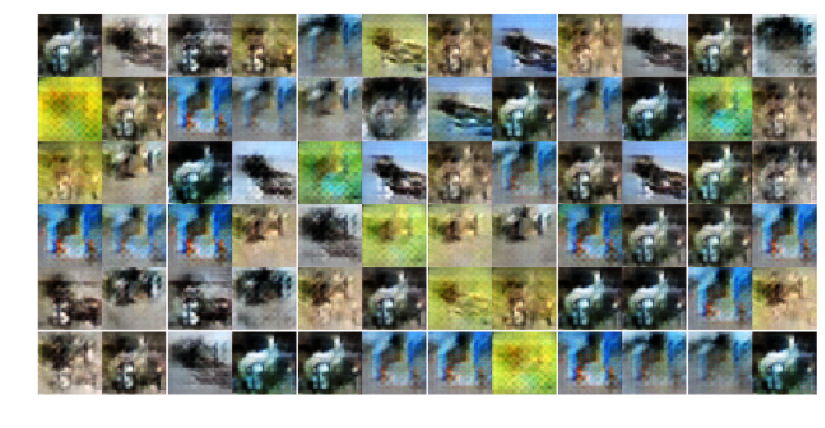}
            \caption{DCGAN 3 (After)}
            \label{fig:dcgan_cifar_after_3}
        \end{subfigure}%
        \caption{Fake Images Generated from DCGAN Before and After Unstable/Mode Collapse for CIFAR-10 dataset}
        \label{fig:10}
\end{figure*}

Further, fake images generated using the DCGAN before and after mode collapse are also shown in Fig. \ref{fig:10}. It can be seen that after mode collapse (\ref{fig:dcgan_cifar_after_1}-\ref{fig:dcgan_cifar_after_3}), the images tend to represent only few of the classes.

\begin{figure}[h]
\centering
        \begin{subfigure}[b]{0.5\textwidth}
            \includegraphics[width=\linewidth]{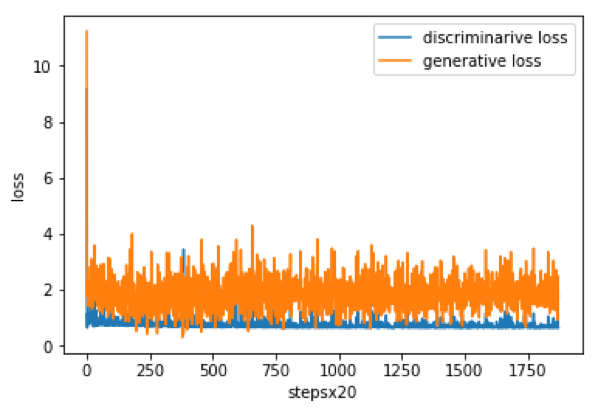}
            \caption{C=0.1}
            \label{fig:lcnn_dcgan_cifar_c_0_1}
        \end{subfigure}%
        \begin{subfigure}[b]{0.5\textwidth}
            \includegraphics[width=\linewidth]{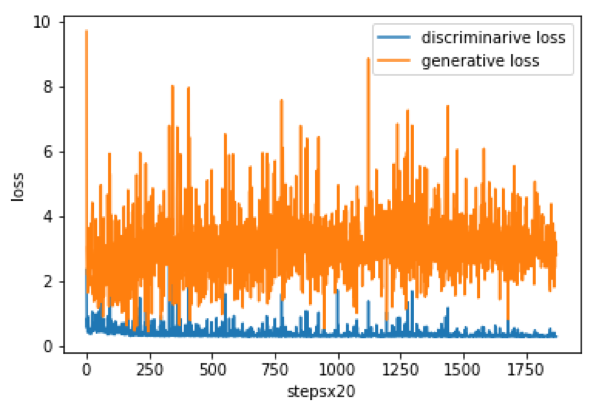}
            \caption{C=0.01}
            \label{fig:lcnn_dcgan_cifar_c_0_01}
        \end{subfigure}%
        \caption{Training Stability for LCNN-DCGAN for different values of C.}
        \label{fig:11}
\end{figure}

We now compare the generative and discriminative loss in case of LCNN-DCGAN in Fig. \ref{fig:11}. The training stability is compared for different values of C and it can be seen that compared to the baseline training stability for DCGAN, there is relatively lower generative loss in case of using the LCNN-DCGAN.

\begin{figure*}
\centering
        \begin{subfigure}[b]{0.29\textwidth}
            \includegraphics[width=\linewidth]{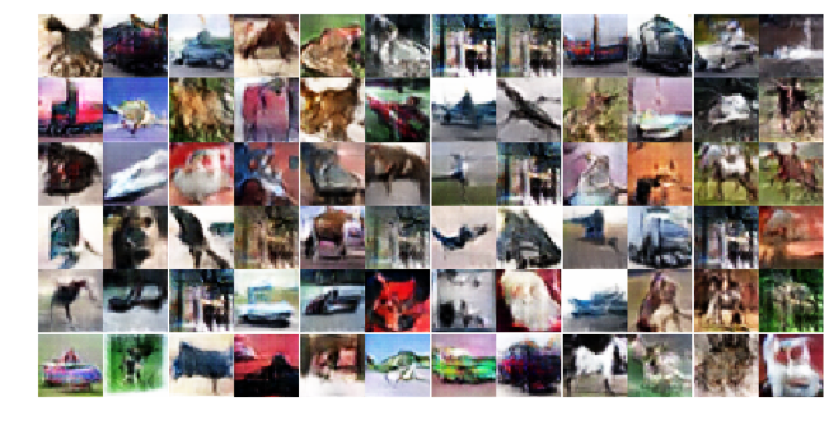}
            \caption{LCNN-DCGAN 1}
            \label{fig:dcgan_lcnn_cifar_1}
        \end{subfigure}%
       \begin{subfigure}[b]{0.29\textwidth}
            \includegraphics[width=\linewidth]{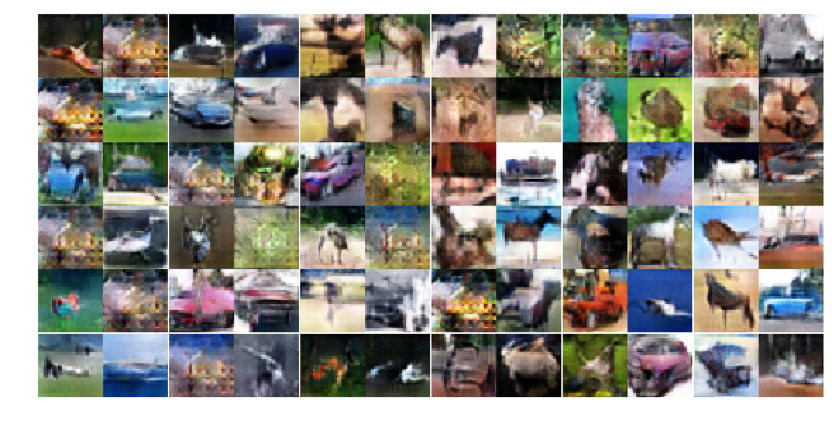}
            \caption{LCNN-DCGAN 2}
            \label{fig:dcgan_lcnn_cifar_2}
        \end{subfigure}%
       \begin{subfigure}[b]{0.29\textwidth}
            \includegraphics[width=\linewidth]{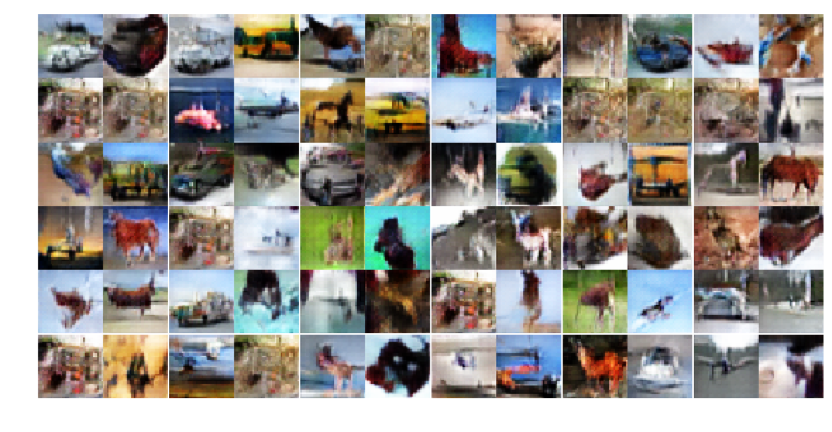}
            \caption{LCNN-DCGAN 3}
            \label{fig:dcgan_lcnn_cifar_3}
        \end{subfigure}%

       \begin{subfigure}[b]{0.29\textwidth}
            \includegraphics[width=\linewidth]{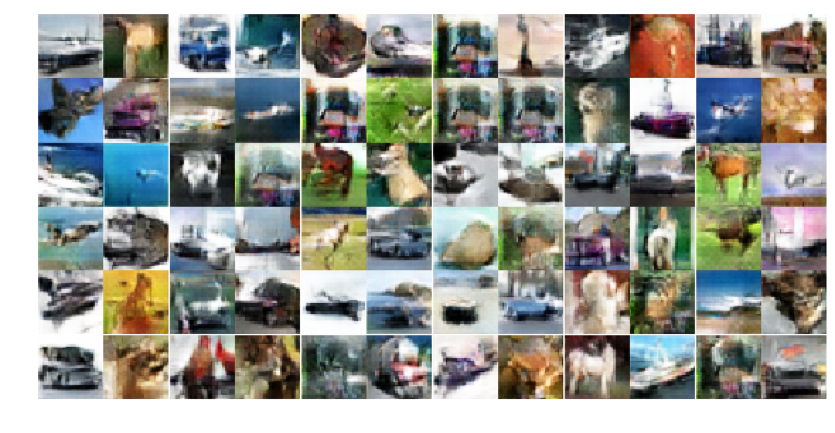}
            \caption{LCNN-DCGAN 4}
            \label{fig:dcgan_lcnn_cifar_4}
        \end{subfigure}%
       \begin{subfigure}[b]{0.29\textwidth}
            \includegraphics[width=\linewidth]{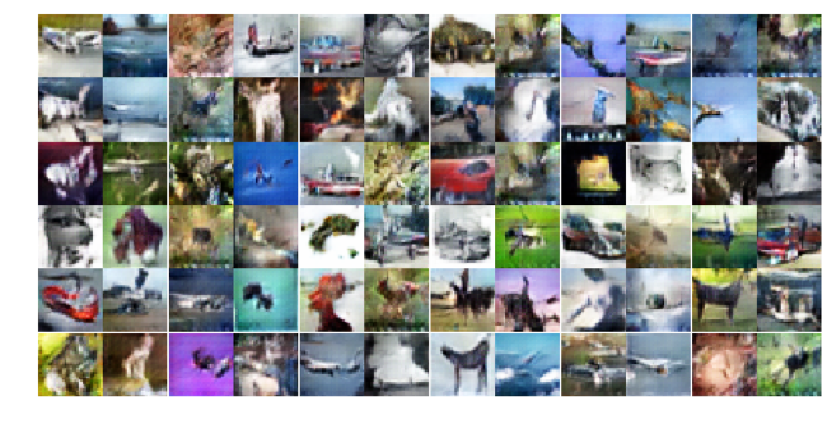}
            \caption{LCNN-DCGAN 5}
            \label{fig:dcgan_lcnn_cifar_5}
        \end{subfigure}%
    \caption{Fake Images Generated from LCNN-DCGAN for CIFAR-10 dataset}
    \label{fig:12}
\end{figure*}

In terms of images from the CIFAR-10 dataset, few representative images are shown in Fig. \ref{fig:12} for the LCNN-DCGAN. It can be seen that mode collapse is avoided when compared to conventional DCGAN.

\subsection{SVHN}

\begin{figure}
    \centering
    \includegraphics[scale=0.8]{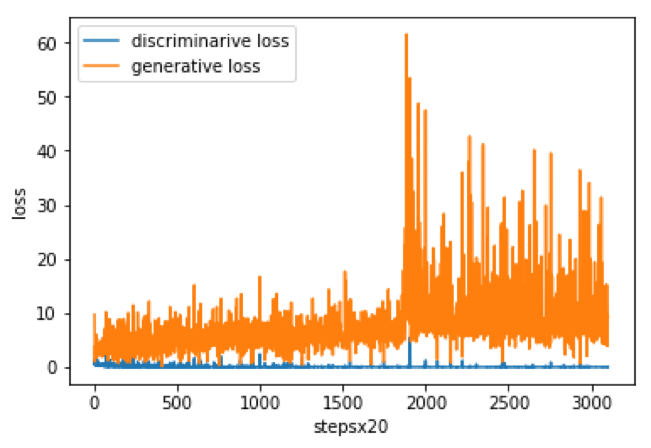}
    \caption{DCGAN: Mode Collapse/Unstable after 1900x20 mini Batches.}
    \label{fig:13}
\end{figure}

\begin{figure*}
\centering
       \begin{subfigure}[b]{0.23\textwidth}
            \includegraphics[width=\linewidth]{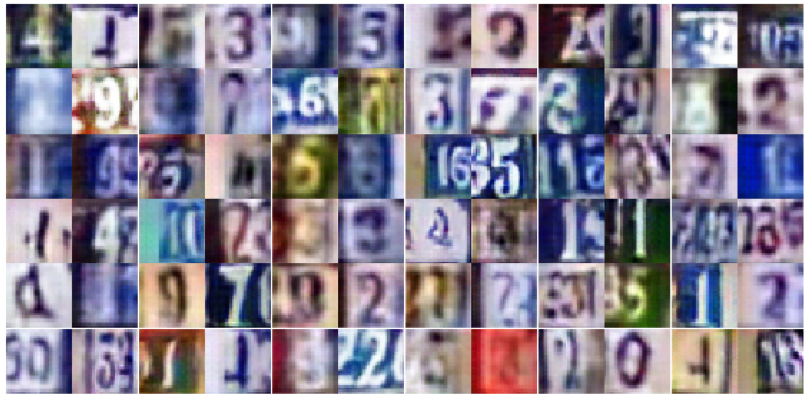}
            \caption{DCGAN (Before 1)}
            \label{fig:dcgan_svhn_before_1}
        \end{subfigure}
       \begin{subfigure}[b]{0.23\textwidth}
            \includegraphics[width=\linewidth]{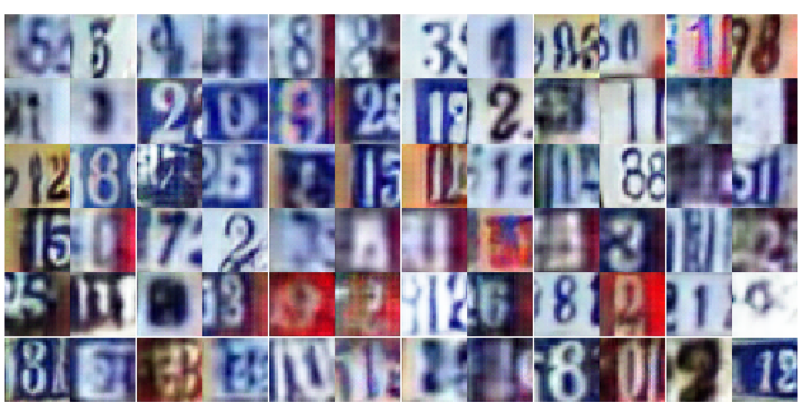}
            \caption{DCGAN (Before 2)}
            \label{fig:dcgan_svhn_before_2}
        \end{subfigure}
       \begin{subfigure}[b]{0.23\textwidth}
            \includegraphics[width=\linewidth]{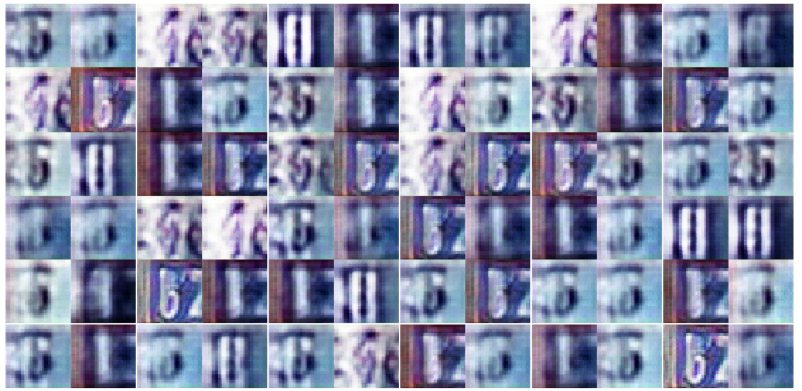}
            \caption{DCGAN (After 1)}
            \label{fig:dcgan_svhn_after_1}
        \end{subfigure}
       \begin{subfigure}[b]{0.23\textwidth}
            \includegraphics[width=\linewidth]{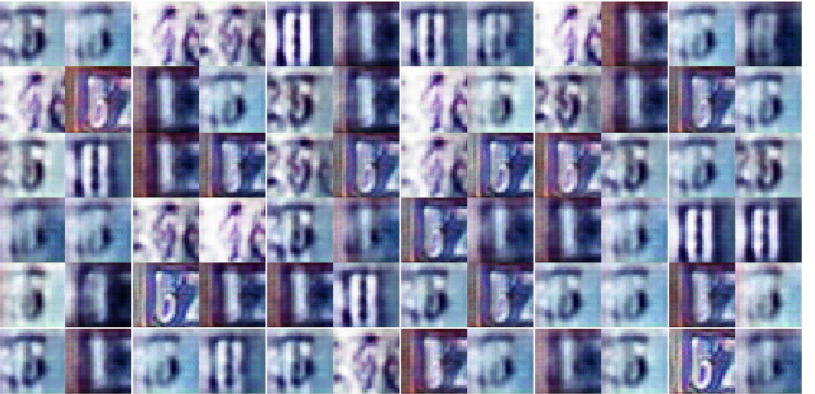}
            \caption{DCGAN (After 2)}
            \label{fig:dcgan_svhn_after_2}
        \end{subfigure}
    \caption{Fake Images Generated from DCGAN Before Unstable/Mode Collapse for SVHN Dataset}
    \label{fig:14}
\end{figure*}

We finally present results on the SVHN dataset. The baseline generative and discriminative loss are shown in Fig. \ref{fig:13} for the DCGAN. Further, representative images for the SVHN dataset are shown in Figs. \ref{fig:dcgan_svhn_before_1}-\ref{fig:dcgan_svhn_before_2} and \ref{fig:dcgan_svhn_after_1}-\ref{fig:dcgan_svhn_after_2} respectively.

\begin{figure}[h]
    \centering
    \includegraphics[scale=0.8]{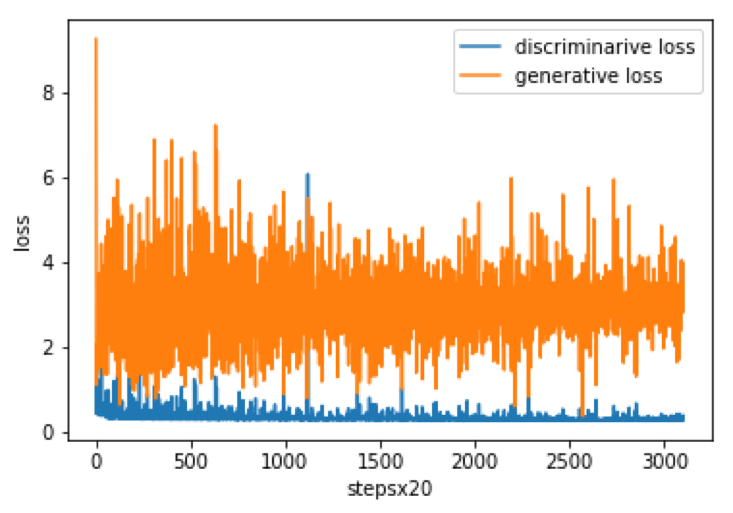}
    \caption{LCNN DCGAN: Stable Training Value of C = 0.01}
    \label{fig:15}
\end{figure}

In case of the LCNN-DCGAN, the training stability for C=0.01 is shown in Fig. \ref{fig:15}. It can be seen that generative loss is significantly contained in case of the LCNN-DCGAN.

\begin{figure}
\centering
        \begin{subfigure}[b]{0.25\textwidth}
            \includegraphics[width=\linewidth]{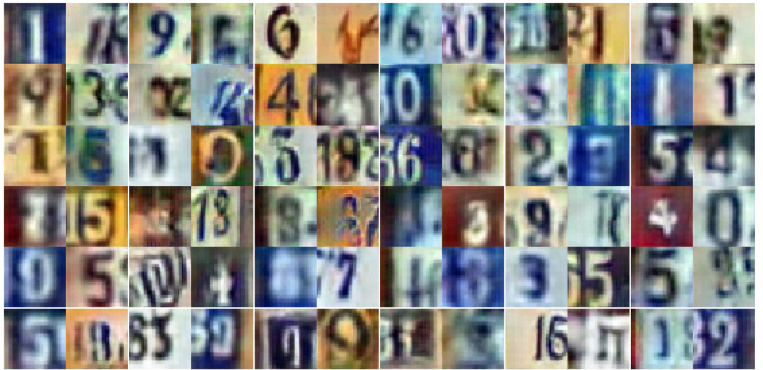}
            \caption{LCNN-DCGAN 1}
            \label{fig:lcnn_dcgan_svhn_1}
        \end{subfigure}
        \begin{subfigure}[b]{0.25\textwidth}
            \includegraphics[width=\linewidth]{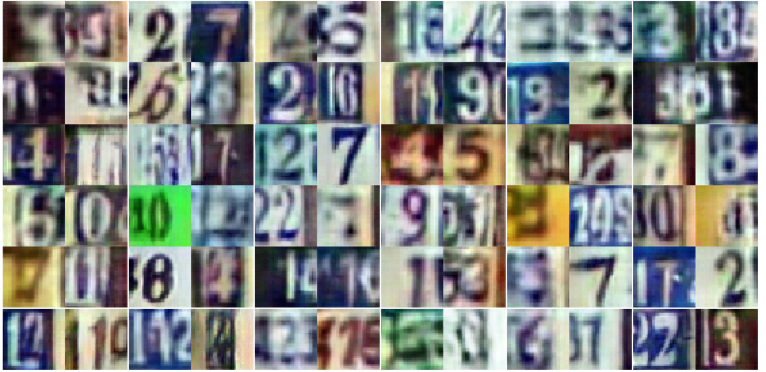}
            \caption{LCNN-DCGAN 2}
            \label{fig:lcnn_dcgan_svhn_2}
        \end{subfigure}
        \begin{subfigure}[b]{0.25\textwidth}
            \includegraphics[width=\linewidth]{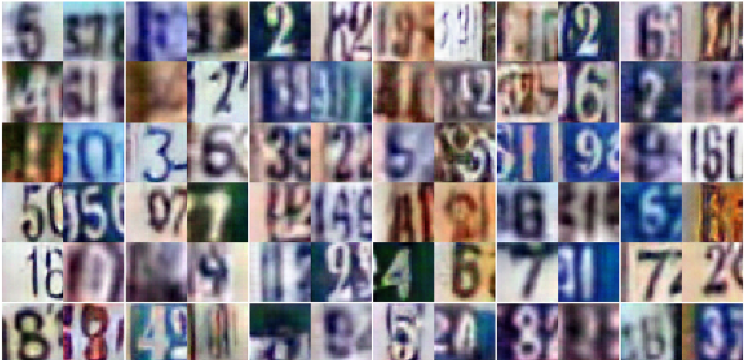}
            \caption{LCNN-DCGAN 3}
            \label{fig:lcnn_dcgan_svhn_3}
        \end{subfigure}%
    \caption{Fake Images Generated from LCNN-DCGAN}
\label{fig:16}
\end{figure}

Finally, representative images from the SVHN dataset for the LCNN-DCGAN are shown in Fig. \ref{fig:16}. It can be seen that mode collapse is avoided when compared to the baseline images obtained using DCGAN.

\subsection{Quantitative Evaluation of LCNN-GAN models}

The inception score is a quantitative metric for determining the performance of GANs. It can be used to determine whether the images have variety and how close these images represent actual images. The score is high when both these factors are true, and low when the converse holds. The Inception Score is based on Google's inception classifier and the lowest possible score is zero. In essence, the inception score compares the label distribution of each image with the entire set of images to determine the difference between these distributions in terms of the Kullback-Leibler (KL) divergence.

Table \ref{tab:inception_score} compares the mean and standard deviation of inception scores for various models for the CIFAR-10 dataset. The inception score has been computed on 5000 generated images which have been split into ten parts and mean and standard deviation for inception scores have been reported. The corresponding learning rate has also been indicated for reference.

\begin{table}[hbtp]
\caption{Results on inception score for CIFAR-10}\label{tab:inception_score}
\centering
\begin{tabular}{|c|c|c|c|c|}
\hline
S.No. & Network Architecture  & Mean & Std. Dev.  & Learning Rate     \\ \hline
1     & DCGAN                 & 5.8  & 0.25 & 0.0005 \\ \hline
2     & DCGAN-LCNN (C=0.01)   & 6.04 & 0.16 & 0.0005 \\ \hline
3     & DCGAN-LCNN (C=0.05)   & 6.33 & 0.2  & 0.0005 \\ \hline
4     & DCGAN-LCNN (C=0.5)    & 6.35 & 0.2  & 0.0005 \\ \hline
5     & DC-SNGAN               & 7.4  & 0.26 & 0.0002 \\ \hline
6     & DC-SNGAN-LCNN (C=0.01) & 7.68 & 0.24 & 0.0002 \\ \hline
7     & DC-SNGAN-LCNN (C=0.05) & 7.63 & 0.21 & 0.0002 \\ \hline
\end{tabular}
\end{table}

\section{Conclusion}
This paper shows that controlling model complexity in GANs significantly alleviates the problem of mode collapse. We use the LCNN model complexity control approach to achieve this objective. Hybrid models obtained by incorporating the LCNN loss function into various GAN models, including the conventional GAN, DC-GAN and SN-GAN yield demonstrable and consistent improvements and do not show mode collapse despite prolonged training. Our approach has shown to provide better control over generator and discriminator losses during training by means of the hyperparameter used in the LCNN functional. The claims have been established by results in terms of modes of generated images, generator/discriminator losses; the inception scores for benchmark datasets show that the use of the LCNN-based loss function avoids mode collapse and provides improved performance across the board for the proposed hybrid models.

\bibliographystyle{unsrt}
\bibliography{ref}

\end{document}